\documentclass{article}

\PassOptionsToPackage{numbers, compress}{natbib}
\usepackage[preprint]{neurips_2026}
\usepackage{subcaption}

\usepackage[utf8]{inputenc} % allow utf-8 input
\usepackage[T1]{fontenc}    % use 8-bit T1 fonts
\usepackage{hyperref}       % hyperlinks
\usepackage{url}            % simple URL typesetting
\usepackage{booktabs}       % professional-quality tables
\usepackage{amsfonts}       % blackboard math symbols
\usepackage{nicefrac}       % compact symbols for 1/2, etc.
\usepackage{microtype}      % microtypography
\usepackage{xcolor}         % colors

\usepackage{graphicx}
\usepackage{arydshln}
\usepackage{amsmath}
\usepackage{amssymb}

\newcommand{\method}{\textsc{BitEmbed}}
\title{BitNet Text Embeddings}

\author{%
  Zhen Li\textsuperscript{1}\thanks{Work done during internship at Microsoft Research.} \quad Xin Huang\textsuperscript{2} \quad Liang Wang\textsuperscript{2} \quad Nan Yang\textsuperscript{2} \quad Ting Song\textsuperscript{2} \quad Yan Xia\textsuperscript{2} \quad Xun Wu\textsuperscript{2} \And \quad Shaohan Huang\textsuperscript{2} \quad Huishuai Zhang\textsuperscript{1} \quad Furu Wei\textsuperscript{2} \quad Dongyan Zhao\textsuperscript{1} \\
  \textsuperscript{1}Peking University \quad \textsuperscript{2}Microsoft Research \\
  \url{https://aka.ms/GeneralAI}
}

\begin{document}

\maketitle
\begin{abstract}
  LLM-based text embedders have substantially improved retrieval and semantic representation quality, but their deployment remains costly: large backbone models slow down embedding inference, while high-dimensional full-precision embeddings impose substantial storage and bandwidth overhead on large-scale indexes. In this paper, we present \method{}, an extreme low-bit framework for LLM-based text embedding that jointly targets encoding efficiency and vector storage. \method{} converts pretrained LLM backbones into BitNet-style embedding encoders with ternary weights, quantized activations, and lightweight normalization refinement. The converted model is adapted to representation learning through continual contrastive pre-training, followed by supervised contrastive fine-tuning with both similarity-distribution distillation and attention-relation distillation from a full-precision teacher. Beyond quantizing the backbone, \method{} further trains output embeddings to support multiple storage precisions meeting different storage needs in various scenarios. Experiments on MMTEB (eng, v2) with Qwen3-0.6B and Gemma3-270M show that \method{} is largely comparable to full precision teacher embedders.
Moreover, \method{} flexibly obtains text embeddings of various precisions, achieving a trade-off between performance and storage cost.
Our model is publicly available at \url{https://huggingface.co/microsoft/bitnet-embedding-0.6b}.
\end{abstract}

\section{Introduction}
\label{sec:introduction}

Text embeddings have become a core interface between natural language and large-scale information systems. Modern retrieval~\cite{nguyen2016ms}, retrieval-augmented generation~\cite{zhao2024retrieval}, recommendation~\cite{zhao2023embedding}, and question answering~\cite{chen2020open} pipelines all rely on embedding models to map text into dense vectors whose geometry reflects semantic relevance~\cite{reimers2019sentence,khattab2020colbert,gao2021simcse,muennighoff2022sgpt}.
Pre-trained bidirectional encoder architectures such as BERT~\cite{devlin-etal-2019-bert} have been widely adopted as backbone models for text embedding~\cite{karpukhin2020dense,xiong2020approximate,zhao2024dense}.
With recent advances in large language models, LLM-based embedders utilize foundation LLMs as the backbone and adapt them with large-scale contrastive learning objectives \citep{li2023towards,su-etal-2023-one,ma2024fine,lee2024gecko,SFRAIResearch2024}. 
With the richer world knowledge and text understanding abilities inherent in LLMs, LLM-based embedders have achieved strong performance across semantic representation tasks.
However, despite these advances, LLM-based embedders still inherit the critical deployment challenges: high embedding inference latency and substantial storage costs.
% In these models, a task instruction and an input text are fed into a decoder-style LLM, and the final hidden state of a special token such as \texttt{<eos>} is used as the fixed-dimensional text representation. 
% This simple formulation has proven highly effective, but it also inherits the deployment cost of LLMs.

Encoding queries and documents with LLM backbones is considerably more expensive than with conventional encoder-only models. Meanwhile, the produced embeddings must be stored, transferred, and searched at massive scale. A production retrieval system may maintain vectors for millions or billions of documents where high-dimensional and high-precision embeddings can dominate memory footprint and data movement even when the encoder itself is optimized. 
These costs become more pronounced as LLM-based embedders are adopted in online retrieval services and storage-constrained scenarios, making efficient inference and representation central requirements for practical deployment.

Extreme low-bit LLMs provide a promising path toward reducing inference cost~\cite{dettmers2022gpt3,li2025quantization,huang2025quaff,gong2025survey,castro2026quartet,wang2026qsvd}. BitNet-style models employ low-precision ternary weights and quantized activations, enabling substantial memory savings and faster inference on suitable hardware while maintaining high precision for the optimizer states and gradients during training~\cite{wang2023bitnet,ma2024era,ma2025bitnet}. BitNet Distillation further shows that full-precision LLMs can be transformed into 1.58-bit task models through architectural stabilization, continued training, and distillation~\cite{wu2025bitnet}. Inspired by these findings, we explore whether similar low-bit principles can be specialized for text embedding.
% , where the central objective is to preserve a semantic similarity space rather than only task labels or token predictions. This perspective motivates an embedding-specific quantization and training framework.

In this paper, we introduce \textbf{BitNet Text Embeddings}, abbreviated as \method{}, a framework for extreme low-bit LLM-based text embedding. \method{} converts pre-trained LLM backbones into BitNet-style embedders, together with normalization modules that stabilize training under quantization. To adapt the quantized backbone to representation learning, \method{} first performs continual contrastive pre-training on large-scale text pairs and then conducts supervised fine-tuning with in-batch and hard negatives. 
Meanwhile, a full-precision fp16 embedding model is tuned as a teacher model, where the quantized embedder is trained not only with contrastive supervision but also to match the teacher's batchwise cosine-similarity distribution and attention relations through distillation.
Beyond quantizing the backbone, \method{} also trains the output embeddings to support multiple storage precisions. Inspired by Matryoshka-style representation and quantization~\cite{kusupati2022matryoshka,tao2024llms,nair2025matryoshka}, we quantize each embedding dimension to multiple precisions during training and optimize the average loss across these precisions. This design encourages a single embedder to produce vectors that remain effective under different memory budgets, allowing practitioners to trade storage for quality without retraining separate models.

We evaluate \method{} on the Massive Multilingual Text Embedding Benchmark (MMTEB)~\cite{enevoldsen2025mmteb}. Our \method{} approaches the performance of full-precision teachers while improving inference efficiency. Furthermore, \method{} also enables the trade-off between storage cost and performance with multi-precision embedding quantization.
% Across vector precisions, 8-bit and 4-bit embeddings retain quality close to FP16, and even 1-bit and 2-bit embeddings preserve useful embedding performance. Ablations show that the proposed training components are complementary and jointly improve the robustness of low-bit text embeddings.

Specifically, our contributions are summarized as follows:
\begin{enumerate}
    \item We present, to the best of our knowledge, the first systematic study of extreme low-bit quantization for LLM-based text embedding, targeting both encoder inference and vector-index storage.
    \item We propose an effective BitNet-style training framework that combines contrastive continual pre-training, supervised fine-tuning, similarity distribution and attention relation distillation.
    % \item We introduce multi-precision embedding training and show that \method{} can approach full-precision teacher performance while supporting efficient 1-, 2-, 4-, 8-, and 16-bit embedding representations.
    \item We introduce multi-precision embedding training and verify that \method{} can approach full-precision teacher performance while supporting efficient multi-precision embedding representations.
\end{enumerate}

\section{Related Work}
\label{sec:related-work}

\subsection{LLM-based Text Embedders}
\label{sec:rw-llm-embedder}
Text embedders are increasingly moving from encoder-only Transformer models~\cite{reimers2019sentence,khattab2020colbert,gao2021simcse,wang2022text,li-etal-2023-faa} toward decoder-only LLM backbones~\cite{muennighoff2022sgpt,li2023towards,wang-etal-2024-improving-text,li2024llama2vec,tao2024llms}. 
% Early explorations treated LLMs as embedding generators typically using last-token or mean pooling over hidden states to obtain text representations~\cite{muennighoff2022sgpt,li2023towards,wang-etal-2024-improving-text,li2024llama2vec,tao2024llms}. 
Recently, LLM-based text embedders have further explored instruction-following, multilinguality, multi-task generalization capabilities~\cite{lee2024gecko,lee2024nv,chen2025mme5,huang2025geogpt}. 
INSTRUCTOR~\cite{su-etal-2023-one} trains embeddings to follow natural-language task instructions, enabling task-specific representations through prompts. 
% BGE-en-ICL~\cite{li2024making} studies few-shot in-context learning for text embedding and shows that task examples can improve embedding quality. 
Recent models also rely on more sophisticated training recipes~\cite{choi2024linq,sorokin2025q,muennighoff2025generative,cai2025revela,zhao2025kalm}. Qwen3-Embedding~\cite{zhang2025qwen3} uses LLMs to synthesize large-scale high-quality data for multi-stage contrastive training. 
% KaLM-Embedding-V2~\cite{zhao2025kalm} incorporates techniques such as focal-style reweighting and online hard-negative mixing. 
These studies focus primarily on improving embedding quality and generalization. In contrast, \method{} studies how to make LLM-based embedders practical under strict inference and storage budgets by combining extreme low-bit quantization with multi-precision output embeddings.

% Text embedding models have evolved from sentence encoders and dense retrievers based on encoder-only Transformers to increasingly capable decoder-only embedders. 
% Early work such as Sentence-BERT, DPR, SimCSE, Contriever and E5 established contrastive learning, dual-encoder retrieval. 
% More recent systems further improve general-purpose embedding quality through larger and more diverse training data, hard-negative mining, multilingual training, and distillation.
% , including BGE-M3, GritLM, GTR, INSTRUCTOR.
% Recent LLM-based embedders use foundation LLMs as representation backbones and often exploit their instruction-following, multilingual, and synthetic-data generation capabilities. Qwen3-Embedding and KaLM-Embedding-V2 demonstrate that multi-stage training, high-quality supervised data, and contrastive distillation can produce strong compact embedders across broad benchmarks \citep{}. Other recent retrieval-oriented work explores richer document representations, such as multi-perspective or multi-vector embeddings, to improve fine-grained query--document matching \citep{}. These studies focus primarily on improving embedding quality and generalization. In contrast, \method{} studies how to make LLM-based embedders practical under strict inference and storage budgets by combining extreme low-bit backbone quantization with multi-precision output embeddings.

\subsection{Quantization for LLMs}
\label{sec:rw-llm-quantization}
Quantization techniques are crucial for deploying Large Language Models (LLMs) under computational and storage constraints.
A common strategy is Post-Training Quantization (PTQ)~\cite{dettmers2022gpt3,yuan2023rptq,xiao2023smoothquant}, encompassing methods like GPTQ~\cite{frantar2022gptq} and AWQ~\cite{lin2024awq}. PTQ is highly efficient as it applies low-bit quantization using only a small set of calibration data. While PTQ methods offer rapid compression, they typically suffer from significant performance degradation at ultra-low bitwidths below 4 bits. To overcome this, Quantization-Aware Training (QAT)~\cite{du2024bitdistiller,liu2024llm,xu2024onebit,chen2025efficientqat} integrates quantization directly into the optimization loop, significantly raising the performance ceiling for low-bit models.
Recently, QAT has been pushed to extreme limits with 1.58-bit (ternary) architectures like BitNet~\cite{wang2023bitnet,ma2024era,ma2025bitnet}. 
Inspired by Bitnet distllation~\cite{wu2025bitnet}, we introduce the 1.58-bit LLM-based text embedder, bridging ternary QAT and representation distillation.
\section{Method}
\label{sec:method}

% \subsection{Overview}
% \label{sec:method-overview}

\method{} is an extreme low-bit LLM-based text embedder which follows a three-stage training pipeline. Firstly, we convert a pretrained LLM into a BitNet-style embedding encoder. Then we adapt the converted encoder with continual contrastive pre-training on large-scale text pairs. Finally we fine-tune the low-bit encoder on supervised training data using contrastive learning with teacher-guided distillation from a FP16 embedding teacher. Additionally, we expose output embeddings to multiple storage precisions, enabling a single \method{} checkpoint to support different memory budgets.

% Starting from a pretrained LLM, we build a BitNet-style embedding encoder whose output embeddings can be stored at multiple precisions.
% \method{} is optimized through continual contrastive pre-training first and then supervised contrastive fine-tuning refines the embedding space with similarity distribution and attention relation distillation by introducing full-precision teacher model.
% The low-bit embedder is optimized with an embedding-oriented training recipe: continual contrastive pre-training first adapts the quantized backbone to representation learning, and supervised contrastive fine-tuning then refines the embedding space with in-batch and hard negatives. In the supervised stage, we further introduce a full-precision embedder as the teacher to provide similarity distribution and attention relation distillation signals. After training, only the low-bit embedder is used for inference and indexing.

\subsection{LLM-based Text Embedding Formulation}
\label{sec:embedding-formulation}

% We use a general formulation that covers common LLM-based embedding architectures. 
Given the input sequence $X=\{x_1,x_2,...,x_n\}$ where $x_i$ represents the $i$-th token in the sequence, the decoder-only LLM $M$ processes this sequence to produce a contextualized representation for each token, denoted as $H(X)=\{h_1, h_2,...,h_n\}$, where $h_i$ is the output hidden state for each token.
A pooling function converts these hidden states into a single representation followed by $\ell_2$ normalization : $h_X=\operatorname{norm}(\operatorname{Pool}(H_{\theta}(X)))$. The pooling function can be last-token pooling~\cite{li2024making}, mean pooling~\cite{muennighoff2025generative}, or architecture-specific strategy~\cite{lee2024nv}. 
In this work, we append an $\text{[EOS]}$ token to each input and use the hidden state of the final token as the text embedding.
% For a query $q$ and a document $d$, their relevance score is the cosine similarity $s_{\theta}(q,d \mid I)=e_{\theta}(I,q)^{\top}e_{\theta}(\emptyset,d)$. In asymmetric retrieval, the instruction is typically attached to the query side; for symmetric tasks, instructions may be attached to both texts.

\subsection{Low-bit Embedding Backbone}
\label{sec:low-bit-backbone}

\method{} is built by applying BitNet-style quantization to the pretrained LLM backbone. Following the BitNet line of work~\cite{wang2023bitnet,ma2025bitnet}, we quantize linear weights in transformer blocks to ternary values. For a full-precision weight matrix $W$, the quantized weight is:
% \begin{small}
\begin{equation}
    Q_w(W)=\Delta_w \cdot
    \operatorname{RoundClip}(\frac{W}{\Delta_w + \epsilon}, -1, 1),
    \label{eq:weight-quant}
\end{equation}
% \end{small}
where $\Delta_w=\operatorname{mean}(|W|)$, $\operatorname{RoundClip}(X,a,b)=\operatorname{max}(a, \operatorname{min}(b, \operatorname{round}(X)))$.
This maps weights to scaled values in $\{-\Delta_w,0,\Delta_w\}$, yielding the 1.58-bit parameterization.

For the activations entering each BitLinear layer, we apply token-wise 8-bit quantization. Given an activation tensor $X$, we compute $\gamma_x=\max |X|$ for each token and apply:
\begin{equation}
    Q_x(X)=
    \frac{\gamma_x}{127}
    \operatorname{RoundClip}(\frac{127}{\gamma_x + \epsilon}X, -128, 127)
    \label{eq:activation-quant}
\end{equation}
% The rounding operations in both weight and activation quantization are optimized with the straight-through estimator (STE). 
The quantizers in Equations~\ref{eq:weight-quant} and~\ref{eq:activation-quant} contain non-differentiable operations, such as rounding and clipping. Following prior low-bit Transformer training practice ~\cite{wang2023bitnet,ma2024era}, we employ the straight-through estimator (STE)~\cite{bengio2013estimating} to approximate the gradients of these quantization operators during back-propagation.

% Since quantization may amplify activation-scale fluctuation, we insert lightweight sub-layer normalization modules inside each transformer block, before the attention output projection and before the feed-forward down projection. This follows the stabilization principle of BitNet-style architectures, but the converted model is trained specifically as a text embedding encoder.

Quantizing both weights and activations makes the transformer more sensitive to activation outliers and scale drift~\cite{ma2024era}. Following prior works~\cite{ma2025bitnet,wu2025bitnet}, we perform model refinement by inserting sub-layer normalization (SubLN) modules inside each transformer block. 
% Standard pre-normalization controls the input to a block, but the intermediate representations before the output projection of the multi-head self attention module and the out projection of the feed-forward network can still have unstable variance after quantization. We therefore normalize these two intermediate states immediately before they are consumed by quantized output projections.
Concretely, let $A_{\ell}$ be the concatenated multi-head attention output before the output projection at layer $\ell$, and let $F_{\ell}$ be the intermediate feed-forward representation before the down projection. We refine the residual updates as
\begin{align}
    Y_{\ell}
    &=
    X_{\ell}
    +
    \operatorname{BitLinear}_{o,\ell}
    \left(
    \operatorname{SubLN}(A_{\ell})
    \right),
    X_{\ell+1}
    =
    Y_{\ell}
    +
    \operatorname{BitLinear}_{d,\ell}
    \left(
    \operatorname{SubLN}(F_{\ell})
    \right),
    \label{eq:subln-ffn}
\end{align}
where $X_{\ell}$ denotes the input hidden states of layer $\ell$, $Y_{\ell}$ denotes the intermediate state after the attention residual update, and $\operatorname{BitLinear}$ denotes a linear projection computed with the quantized weights and activations above. By normalizing these intermediate states locally, the quantized projections receive inputs with more consistent scales, which makes optimization less sensitive to low-bit numerical noise.
% For embedding models, this stabilization also helps preserve the geometry of the final representation space, since unstable intermediate activations can propagate to the pooled text vector and distort cosine similarities.

% \subsection{Distillation-aware Training}
% \label{sec:distillation-aware-training}

% The training objective is designed around embedding geometry. We first adapt the low-bit embedder with continual contrastive pre-training, and then conduct supervised fine-tuning with distillation. The full-precision teacher used in the distillation losses is trained beforehand from the original LLM backbone on supervised embedding data using contrastive learning, and remains fixed when supervising the low-bit model.

\subsection{Continual Contrastive Pre-training}
\label{sec:continual-pretraining}

% After BitNet conversion, the low-bit embedder needs to adapt its quantized parameterization to representation learning. We therefore continue training it on large-scale text pairs before supervised fine-tuning. Given a mini-batch $\mathcal{B}=\{(I_i,q_i,p_i^+)\}_{i=1}^{N}$, where $p_i^+$ is a positive text for query $q_i$, we use in-batch positives from other instances as negatives:
% After BitNet conversion, the low-bit embedder can no longer fully preserve the knowledge encoded in the full-precision LLM parameters. 
Extreme quantization can distort the pretrained representation space and weaken the semantic knowledge encoded in the original LLM.
With such an extreme quantization constraint, relying only on supervised fine-tuning data is insufficient for the model to rebuild a reliable semantic representation. We therefore introduce continual contrastive pre-training on large-scale text pairs, which exposes the quantized model to broad semantic co-occurrence and relevance signals before supervised training on the downstream task. Given a training batch $\mathcal{B}=\{(q_i,p_i^+)\}_{i=1}^{N}$, where $p_i^+$ is a positive text for query $q_i$, 
we feed them into the embedder to obtain embedding $h_{q_i}$ and $h_{p_i^+}$ respectively.
% we add special token $\texttt{[EOS]}$ to end of the input $q_i$ and $p_i^+$ and feed them to embedder to obtain embedding $h_{q_i}$ and $h_{p_i^+}$ respectively by last-token pooling.
Then we employ the standard InfoNCE~\cite{oord2018representation} loss for optimization utilizing in-batch negatives:

\begin{equation}
    \mathcal{L}_{\mathrm{cpt}}
    =
    -\frac{1}{N}
    \sum_{i=1}^{N}
    \log
    \frac{\exp(s_{\theta}(h_{q_i},h_{p_i^+})/\tau)}
    {\sum_{j=1}^{N}\exp(s_{\theta}(h_{q_i},h_{p_j^+})/\tau)}.
    \label{eq:cpt-loss}
\end{equation}
where $s(·)$ measures the cosine similarity between two embedding vectors and $\tau$ is the temperature coefficient.
% This stage serves as a warm-up that lets the low-bit encoder form a preliminary semantic space before it is exposed to high-quality supervised data.
This stage provides broad contrastive adaptation and initializes a semantic embedding space before supervised fine-tuning.

\subsection{Distillation-based Supervised Fine-tuning}
\label{sec:distillation-aware-training}

After continual pre-training, we train the low-bit embedder on high-quality supervised data. 
A fine-tuning batch is organized as $\{(I_i,q_i,p_i^+,\{p_{i,k}^-\}_{k=1}^{K})\}_{i=1}^{N}$, where the query $q_i$ is formatted with its task instruction $I_i$ and $p_i^+$ and $\{p_{i,k}^-\}_{k=1}^{K}$ represent the positive passage and $K$ hard negatives.
% Following common LLM-based embedding recipes \citep{}, each instance contains a query, a positive passage, and $K$ hard negatives, denoted as $(q_i,p_i^+,\{p_{i,k}^-\}_{k=1}^{K})$. The query $q_i$ is formatted with its task definition when available. 
We optimize an InfoNCE loss with both in-batch and mined hard negatives. For each query $q_i$, the candidate set is $\mathcal{C}_i = \{p_j^+\}_{j=1}^{N}
    \cup
    \{p_{i,k}^- , 1\le k\le K\}_{i=1}^{N}$.
% \begin{equation}
%     \mathcal{C}_i =
%     \{p_j^+\}_{j=1}^{N}
%     \cup
%     \{p_{i,k}^- , 1\le k\le K\}.
%     \label{eq:candidate-set}
% \end{equation}
Given this candidate set, the supervised contrastive loss is
\begin{equation}
    \mathcal{L}_{\mathrm{ctr}}
    =
    -\frac{1}{N}
    \sum_{i=1}^{N}
    \log
    \frac{
        \exp(s_{\theta}(h_{q_i},h_{p_i^+})/\tau)
    }{
        \sum_{c\in\mathcal{C}_i}
        \exp(s_{\theta}(h_{q_i},h_{c})/\tau)
    }.
    \label{eq:contrastive-loss}
\end{equation}
% This objective pulls matched query-passage pairs together while contrasting them with other passages in the batch and mined hard negatives, refining the semantic space learned during continual pre-training. 
To further compensate for the representation loss introduced by extreme low-bit quantization, we augment supervised fine-tuning with teacher-guided distillation.
The full-precision teacher model is trained beforehand from the original LLM backbone on the supervised fine-tuning data, and is then fixed to provide distillation signals. We introduce two complementary distillation losses on top of the supervised contrastive objective.

% After continual pre-training, we train the low-bit embedder on high-quality supervised pairs and apply teacher-guided distillation. The full-precision teacher model is trained beforehand from the original LLM backbone on supervised fine-tuning data with contrastive learning. 
% In this stage, each query is formatted with its task definition when available, and the model is optimized with hard contrastive supervision, soft relevance distributions, and internal attention relations from the teacher.

% \paragraph{Contrastive fine-tuning.}
% In the supervised stage, each instance contains a query, a positive passage, and $K$ hard negatives, denoted as $(q_i,p_i^+,\{p_{i,k}^-\}_{k=1}^{K})$. Following common LLM-based embedding recipes \citep{}, we train with both in-batch negatives and hard negatives. For each query $q_i$, the candidate set is
% \begin{equation}
%     \mathcal{C}_i =
%     \{p_j^+\}_{j=1}^{N}
%     \cup
%     \{p_{j,k}^- \mid 1\le j\le N, 1\le k\le K\}.
%     \label{eq:candidate-set}
% \end{equation}
% The supervised contrastive loss is
% \begin{equation}
%     \mathcal{L}_{\mathrm{ctr}}
%     =
%     -\frac{1}{N}
%     \sum_{i=1}^{N}
%     \log
%     \frac{
%         \exp(s_{\theta}(q_i,p_i^+)/\tau)
%     }{
%         \sum_{c\in\mathcal{C}_i}
%         \exp(s_{\theta}(q_i,c)/\tau)
%     }.
%     \label{eq:contrastive-loss}
% \end{equation}
% When duplicate positives or near-duplicate passages appear in the same batch, we mask them out from the denominator to reduce false-negative noise.

\paragraph{Similarity-distribution distillation.}
In addition to the hard supervision used in the contrastive loss, we distill the teacher’s soft similarity distribution over the candidate set. This distribution provides fine-grained relative preference signals among positives and negatives candidates encoded by the teacher.
% Instead of training solely with coarse-grained hard supervised labels, we further perform contrastive distillation by distilling fine-grained soft signals which is the normalized distribution of cosine similarity scores from the teacher model.
% Hard labels alone specify which passage is positive, but they do not describe how the teacher ranks all candidates in the batch. We therefore distill the teacher's score distribution. 
For $M\in\{T,S\}$, where $T$ is the teacher and $S$ is the low-bit embedder, define
\begin{equation}
    P_M(c\mid q_i)=
    \frac{
    \exp(s_M(q_i,c)/\tau)
    }{
    \sum_{c'\in\mathcal{C}_i}
    \exp(s_M(q_i,c')/\tau)
    }.
    \label{eq:score-distribution}
\end{equation}
The score-level distillation loss minimizes the KL divergence from the teacher distribution to the low-bit student model distribution:
\begin{equation}
    \mathcal{L}_{\mathrm{score}}
    =
    \frac{1}{N}
    \sum_{i=1}^{N}
    D_{\mathrm{KL}}
    \left(
    P_T(\cdot\mid q_i)
    \parallel
    P_S(\cdot\mid q_i)
    \right).
    \label{eq:score-distill}
\end{equation}
% This objective is embedding-specific: it transfers the teacher's relative similarity structure, which is exactly the signal used for nearest-neighbor retrieval.
This encourages the embedding model to capture nuanced differences between the positive and negative learning from teacher model.

\paragraph{Attention relation distillation.}
Score distillation aligns the final embedding similarities, but it does not directly constrain how the low-bit encoder organizes token-level interactions before pooling. We therefore adopt a multi-head attention distillation objective inspired by MiniLM~\cite{wang2020minilm,wang2021minilmv2}. 
% Instead of matching raw attention weights only, we distill relation distributions derived from the projected query, key, and value states.

For a selected set of layers $\Omega$ and projection types $\Phi=\{Q,K,V\}$, let $A_{\ell,\phi,a}^{M}\in\mathbb{R}^{L\times d_h}$ denote the normalized projected states of head $a$ from model $M\in\{T,S\}$, where $L$ is the sequence length and $d_h$ is the hidden dimension. $T$ is the full-precision teacher, $S$ is the low-bit embedder, and $N_h$ is the number of heads. For token position $t \in [1, L]$, we compute its relation distribution to all positions as
\begin{equation}
    R_{\ell,\phi,a,t}^{M}
    =
    \operatorname{softmax}
    \left(
    \frac{
    A_{\ell,\phi,a,t}^{M}
    (A_{\ell,\phi,a,t}^{M})^{\top}
    }{\sqrt{d_h}}
    \right),
    \label{eq:attention-relation}
\end{equation}
% where the softmax is applied over the sequence dimension and $\tau_a$ is a temperature. 
The attention distillation loss averages the KL divergence between teacher and low-bit relation distributions:
\begin{align}
    \mathcal{L}_{\mathrm{attn}}
    =
    \frac{1}{|\Omega||\Phi|}
    \sum_{\ell\in\Omega}
    \sum_{\phi\in\Phi}
    \frac{1}{N_hL}
    \sum_{a=1}^{N_h}
    \sum_{t=1}^{L}
    D_{\mathrm{KL}}\left(
    R_{\ell,\phi,a,t}^{T}
    \parallel
    R_{\ell,\phi,a,t}^{S}
    \right).
    \label{eq:attention-distill}
\end{align}
Following MiniLM~\cite{wang2020minilm,wang2021minilmv2}, we perform attention relation distillation at only one layer.
This distillation objective encourages the low-bit student model to retain the teacher's fine-grained structural dependencies.
% , which complements the batch-level similarity distillation used for the final embeddings.

Finally, the distillation-based supervised objective is: $\mathcal{L}_{\mathrm{distill}}
    =
    \mathcal{L}_{\mathrm{ctr}}
    +
    \lambda_s \mathcal{L}_{\mathrm{score}}
    +
    \lambda_a \mathcal{L}_{\mathrm{attn}},$
% \begin{equation}
%     \mathcal{L}_{\mathrm{distill}}
%     =
%     \mathcal{L}_{\mathrm{ctr}}
%     +
%     \lambda_s \mathcal{L}_{\mathrm{score}}
%     +
%     \lambda_a \mathcal{L}_{\mathrm{attn}},
%     \label{eq:distill-loss}
% \end{equation}
where $\lambda_s$ and $\lambda_a$ are hyperparameters which balance score-level and attention-level distillation.

\subsection{Multi-precision Embedding Quantization}
\label{sec:embedding-quantization}
In addition to quantizing the backbone, we train the output embeddings to remain robust under multiple storage precisions during the supervised fine-tuning, which enables deployment under different memory budgets.
% In addition to quantizing the backbone, we train the output embeddings to be robust under multiple storage precisions which meets the needs of storage constrained scenarios.
Let $e\in\mathbb{R}^{d}$ denote the fp16 embedding produced by \method{}, and let $\mathcal{R}=\{1,2,4,8,16\}$denote the set of supported output bitwidths. We use per-vector absmax scaling, $\alpha(e)=\max_j |e_j|$, with a small lower bound for numerical stability. The 16-bit branch returns the original embedding unchanged, while lower-bit settings store compact codes and use their reconstructed values for similarity computation during training.
For 1-bit embeddings, we use scaled binary quantization:
\begin{gather}
    z^{(1)}_j=\operatorname{sgn}_{+}(e_j)\in\{-1,+1\}
    % \quad
    \label{eq:binary-embedding-quant}
\end{gather}
where $\tilde{e}^{(1)}_j=\alpha(e)z^{(1)}_j, \operatorname{sgn}_{+}(0)=1$.
For $b\in\{2,4,8\}$, we use all $2^b$ uniformly spaced levels in $[-\alpha(e),\alpha(e)]$. Let $m_b=2^b-1$. Each dimension is quantized and dequantized as
\begin{gather}
    z^{(b)}_j
    =
    \operatorname{clip}
    \left(
    \operatorname{round}
    \left(
    \frac{e_j/\alpha(e)+1}{2}m_b
    \right),
    0,m_b
    \right)
    \label{eq:uniform-embedding-quant}
\end{gather}
% For example, 2-bit quantization yields four levels $\{-1,-1/3,1/3,1\}\alpha(e)$. 
where $\tilde{e}^{(b)}_j
    =
    \left(
    \frac{2z^{(b)}_j}{m_b}-1
    \right)\alpha(e)$.
Here $z^{(b)}$ is the discrete code stored together with the scale $\alpha(e)$, while $\tilde{e}^{(b)}$ is the reconstructed vector used to compute similarities. 
% During training and evaluation, $\tilde{e}^{(b)}$ is L2-normalized
We use STE during training.
% so that gradients pass through the embedding quantizer.

Inspired by Matryoshka-style multi-precision training~\cite{nair2025matryoshka}, we optimize the \method{} across all supported output embedding precisions. 
% For each $b$, low-bit similarities in the contrastive and score-distillation losses are computed, while the teacher distribution remains full precision. 
The final multi-precision objective is
\begin{equation}
    \mathcal{L}_{\mathrm{mp}}
    =
    \frac{1}{|\mathcal{R}|}
    \sum_{b\in\mathcal{R}}
    \left(
    \mathcal{L}_{\mathrm{ctr}}^{(b)}
    +
    \lambda_s \mathcal{L}_{\mathrm{score}}^{(b)}
    \right) + \lambda_a \mathcal{L}_{\mathrm{attn}}.
    \label{eq:multi-precision-loss}
\end{equation}
% Unlike weight-level Matryoshka quantization, this multi-precision training is applied to the output representation. The embedding dimensionality is unchanged, but the stored precision can be selected according to the deployment budget.

% \subsection{Training and Inference}
% \label{sec:training-inference}

% Training proceeds as follows. We first train the full-precision teacher by supervised contrastive learning from the original LLM backbone. We then construct the low-bit embedder by applying BitNet-style conversion to the LLM backbone, perform continual contrastive pre-training with Equation~\ref{eq:cpt-loss}, and finally optimize the supervised multi-precision distillation objective in Equation~\ref{eq:multi-precision-loss}. The teacher is fixed during distillation and is not used at inference time.

% During inference, the low-bit text embedder encodes an input text once and outputs an embedding at the requested precision. Since all precision levels share the same dimensionality, the vectors can be used by standard dense retrieval systems without modifying the downstream interface. Higher precisions favor quality, while 4-bit, 2-bit, and 1-bit embeddings provide increasingly compact indexes for storage-constrained scenarios.

\section{Experiments}
\label{sec:experiments}

\subsection{Experimental Setup}
\label{sec:exp-setup}

\paragraph{Backbones.}
We evaluate two LLM backbones: Qwen3-0.6B~\cite{qwen3} and Gemma3-270M~\cite{gemmateam2025gemma3technicalreport} which differ in scales and architectures, to examine whether extreme low-bit training can preserve the representation quality across different LLM-based embedders. 
% For each backbone, we first train a full-precision teacher embedder based on the original LLM using supervised contrastive learning. We then construct the low-bit embedder through BitNet-style conversion, continual contrastive pre-training, and distillation-based supervised fine-tuning.

\begin{table*}[t]
\centering
\small
\caption{Main results on MMTEB (eng, v2). We compare \method{} with full-precision teacher. Cls., Clust., PairCls., Rerank, Retr., STS, Summ., and Avg. denote classification, clustering, pair classification, reranking, retrieval, semantic textual similarity, summarization, and the overall average across all tasks, respectively. ``speed" denotes the token throughput of models on CPU with 8 threads.}
\resizebox{\textwidth}{!}{
\begin{tabular}{lcccccccccc}
\toprule
\textbf{Model} & \textbf{Cls.} & \textbf{Clust.} & \textbf{PairCls.} & \textbf{Rerank} & \textbf{Retr.} & \textbf{STS} & \textbf{Summ.} & \textbf{Avg.} & \textbf{Speed}\\
\midrule
\multicolumn{9}{l}{\textbf{\textit{Backbone: Qwen3-0.6B}}} \\
FP16 teacher & 86.37 & 55.48 & 82.56 & 43.89 & 55.34 & 81.15 & 31.87 & 67.95 & 364.36 \\ \hdashline[1pt/1pt]
% \method{}-1bit  & 85.54 & 53.00 & 80.67 & 39.66 & 46.06 & 80.16 & 27.91 & 64.38 \\
% \method{}-2bit  & 85.71 & 53.07 & 80.93 & 40.52 & 47.33 & 80.27 & 27.10 & 64.81 \\
% \method{}-4bit  & 86.15 & 54.74 & 82.21 & 41.97 & 53.87 & 81.32 & 30.08 & 67.28 \\
% \method{}-8bit  & 86.19 & 54.83 & 82.33 & 42.35 & 54.21 & 81.33 & 31.34 & 67.45 \\
\method{} & 86.49 & 55.42 & 82.30 & 43.41 & 54.03 & 81.15 & 32.06 & 67.60 & \textbf{830.50}\\
\midrule
\multicolumn{9}{l}{\textbf{\textit{Backbone: Gemma3-270M}}} \\
FP16 teacher & 86.54 & 53.85 & 80.67 & 43.38 & 52.97 & 80.57 & 28.36 & 66.71 & 1181.28 \\ \hdashline[1pt/1pt]
% \method{}-1bit  & 83.83 & 51.15 & 79.36 & 40.55 & 42.53 & 81.05 & 29.84 & 63.01 \\
% \method{}-2bit  & 83.91 & 51.25 & 79.64 & 41.19 & 43.22 & 80.88 & 30.30 & 63.24 \\
% \method{}-4bit  & 84.99 & 52.23 & 80.91 & 42.17 & 51.11 & 81.16 & 32.55 & 65.96 \\
% \method{}-8bit  & 85.26 & 53.07 & 80.47 & 42.25 & 51.50 & 81.09 & 32.96 & 66.11 \\
\method{} & 85.25 & 53.07 & 80.47 & 42.14 & 51.47 & 81.09 & 32.97 & 66.10 & \textbf{2055.47} \\
\bottomrule
\end{tabular}}
\label{tab:main-results}
\end{table*}

\paragraph{Training data.}
% \paragraph{Continual Pre-training.}
% The continual pre-training stage uses large-scale text-pair data in the same contrastive learning format.
For continual pre-training, following Multilingual E5~\cite{wang2024multilingual}, we train \method{} on 1B text pairs.
For supervised fine-tuning, we train \method{} on the publicly available training data provided by BGE-en-ICL~\cite{li2024making}. More details on training data are available in Appendix~\ref{appendix: training_data}.
Specifically, during supervised fine-tuning, the training data are organized in a retrieval-style format, where each instance contains a query, a corresponding task instruction, a positive passage, and mined hard negatives.
To enable instruction-aware embedding learning, we prepend the task instruction to the query side following the instruction-aware training protocol commonly used by LLM-based embedders:
\begin{align}
    q_{inst} = \texttt{⟨Instruct⟩}\ \{\texttt{task\ definition}\}\ \texttt{⟨query⟩}\ \{q\}
\end{align}

\paragraph{Evaluation.}
We evaluate \method{} on Massive Multilingual Text Embedding Benchmark (MMTEB) (eng, v2)~\cite{enevoldsen2025mmteb}, which is an optimized version of MTEB~\cite{muennighoff-etal-2023-mteb}. 
MMTEB covers diverse embedding tasks, including classification, clustering, pair classification, reranking, retrieval, semantic textual similarity (STS), and summarization.
Following the official metrics, we use accuracy for classification, V-measure for clustering, average precision for pair classification, mean average precision (MAP) for reranking, nDCG@10 for retrieval, and Spearman correlation for STS and summarization.
Meanwhile we also evaluate the runtime efficiency of models which reports the token throughput (tokens per second) on CPU with 8 threads.
The evaluation instructions are provided in Appendix~\ref{appendix: evaluation}.
% We report scores for each task type and the average across all tasks. 
Since our goal is to study performance-preserving quantization for LLM-based text embedding, we compare our \method{} with an FP16 model based on the same backbone which is fine-tuned directly on the same supervised fine-tuning data, rather than comparing against unrelated large-scale embedding systems.
% we use the FP16 model with the same backbone as the primary reference rather than comparing against unrelated large-scale embedding systems.

\paragraph{Implementation details.}
We use a batch size of 128 for supervised fine-tuning, and set $\tau$ in Eq~\ref{eq:cpt-loss}, ~\ref{eq:contrastive-loss} and ~\ref{eq:score-distribution} as 0.02. The number of hard negatives in supervised fine-tuning data is set as 7.
For each backbone, the full-precision teacher is trained on the supervised data using a learning rate of 5e-6, 
% The \method{} is initialized from the same pretrained backbone after BitNet-style conversion. We first perform continual contrastive pre-training on large-scale text pairs, and then conduct supervised fine-tuning on the same data used for teacher training, using in-batch negatives and mined hard negatives. During supervised low-bit training, the teacher is fixed and only provides distillation targets.
and for \method{}, the learning rate of distillation-based supervised fine-tuning is set as 3e-5.
% During distillation-based supervised fine-tuning, the low-bit \method{} is trained with a learning rate of 3e-5. 
The coefficient of similarity-distribution distillation is set to $\lambda_s=0.2$, and the coefficient of multi-head attention distillation is set to $\lambda_a=10^4$. Following MiniLM-style distillation~\cite{wang2020minilm}, we distill attention relations from a single transformer layer, where we use layer 18 in the Qwen3-0.6B and layer 10 in the Gemma3 270M. 
% During multi-precision embedding training, we optimize over $\{1,2,4,8,16\}$-bit output embeddings by averaging the corresponding losses. 

% At evaluation, \method{}-$b$bit denotes the same trained low-bit encoder whose output embeddings are stored and scored at $b$-bit precision; the 16-bit branch uses the unquantized output vector.

% The full-precision teacher is trained with a learning rate of $5\times10^{-6}$, and the low-bit embedder is trained with a learning rate of $3\times10^{-5}$. For distillation, the coefficient of similarity-distribution distillation is set to $\lambda_s=0.2$, and the coefficient of multi-head attention distillation is set to $\lambda_a=10^4$. Following MiniLM-style distillation, we distill attention relations from a single layer rather than all layers; unless otherwise specified, we use layer 22. During multi-precision embedding training, we optimize over $\{1,2,4,8,16\}$-bit output embeddings.

\subsection{Main Results}
\label{sec:main-results}

\paragraph{Overall performance.}
We compare \method{} with the full precision teacher baselines on Qwen3-0.6B and Gemma3-270M, and report the average score for each MMTEB (eng, v2) task category and the overall average score across all tasks in Table~\ref{tab:main-results}. 
The detailed results on individual tasks are provided in the Appendix~\ref{appendix: overall_mmteb}.
Furthermore, for the multilingual scenarios, we also conduct experiments and evaluate on multilingual MMTEB. The experiment details are provided in Appendix~\ref{appendix: multilingual_mmteb}.
Table~\ref{tab:main-results} shows that despite using an extreme low-bit embedding backbone, \method{} remains largely comparable to the full-precision teacher across both backbones. On Qwen3-0.6B, \method{} achieves 67.60, only 0.35 points lower than the FP16 teacher. On Gemma3-270M, \method{} obtains 66.10, closely matching the teacher score of 66.71. These small gaps indicate that BitNet-style backbone conversion, when coupled with contrastive adaptation and teacher-guided distillation, can recover most of the semantic representation quality required by text embedding tasks. 

Meanwhile, \method{} substantially improves inference efficiency. On CPU with 8 threads, BITEMBED improves token throughput from 364.36 to 830.50 tokens/s on Qwen3-0.6B and from 1181.28 to 2055.47 tokens/s on Gemma3-270M, achieving approximately 2× higher CPU token throughput on both backbones. These results show that the practical utility of \method{} in computationally limited scenarios. 

% delivers a significant improvement in system efficiency, achieving approximately 2× higher CPU token throughput on both backbones, which demonstrates the practical utility of \method{} in computationally limited scenarios.
% Moreover, the same pattern holds for two different LLM backbones: Qwen3-0.6B and Gemma3-270M differ in model scale and architecture, yet \method{} consistently tracks the performance of the corresponding full-precision baseline. This suggests that the proposed framework is not specialized to a single model family, but can serve as a general recipe for constructing low-bit LLM-based embedders under different deployment budgets.

\begin{figure}[t]
\centering
\includegraphics[width=0.7\textwidth]{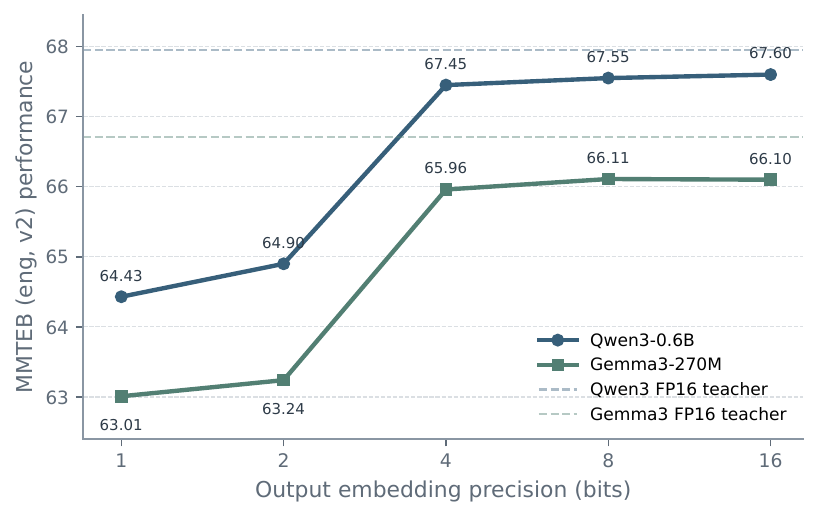}
\caption{Performance-precision trade-off of \method{} on Qwen3-0.6B and Gemma3-270M. We report the average MMTEB (eng, v2) performance of 1-, 2-, 4-, 8-, and 16-bit output embeddings of \method{}.}
\label{fig:precision-bar}
\end{figure}

\begin{table*}[t]
\centering
\small
\caption{Ablation study on Qwen3-0.6B with 16-bit output embeddings. CPT, SD, and AD denote continual pre-training, similarity-distribution distillation, and attention-relation distillation, respectively. BitNet SFT denotes supervised fine-tuning after BitNet conversion, without continual pre-training and distillation.}
\resizebox{\textwidth}{!}{
\begin{tabular}{lcccccccc}
\toprule
\textbf{Variant} & \textbf{Cls.} & \textbf{Clust.} & \textbf{PairCls.} & \textbf{Rerank} & \textbf{Retr.} & \textbf{STS} & \textbf{Summ.} & \textbf{Avg.} \\
\midrule
\method{} & 86.49 & 55.42 & 82.30 & 43.41 & 54.03 & 81.15 & 32.06 & 67.60 \\ \hdashline[1pt/1pt]
w/o. SubLN & 85.31 & 52.80 & 79.90 & 41.16 & 51.01 & 79.57 & 31.34 & 65.48 \\
w/o. AD & 85.70 & 53.21 & 82.56 & 41.85 & 52.89 & 80.50 & 30.19 & 66.49 \\
w/o. SD & 86.12 & 52.95 & 82.53 & 42.23 & 51.40 & 81.40 & 29.97 & 66.37 \\
w/o. SD \& AD & 85.17 & 52.20 & 79.48 & 42.23 & 49.42 & 78.88 & 29.07 & 64.76 \\
w/o. CPT & 84.99 & 52.84 & 79.02 & 41.31 & 48.00 & 80.26 & 30.84 & 64.77 \\
BitNet SFT & 77.72 & 48.79 & 72.77 & 40.39 & 39.73 & 75.35 & 29.13 & 58.92 \\
\bottomrule
\end{tabular}}
\label{tab:ablation}
\end{table*}

\begin{figure}[t]
\centering
\includegraphics[width=\linewidth]{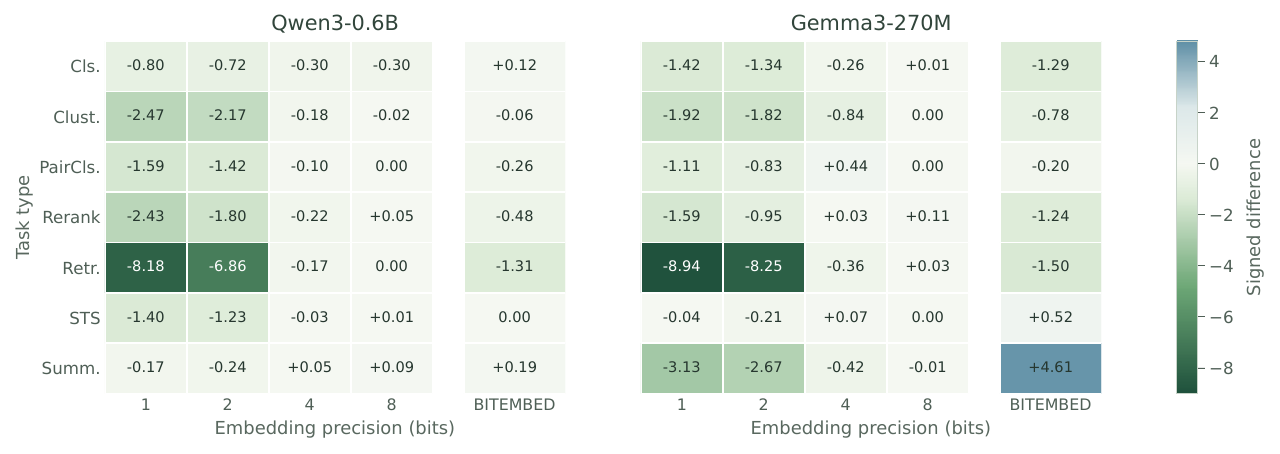}
\caption{Task-type sensitivity on MMTEB (eng, v2). Columns 1, 2, 4, and 8 report the performance differences relative to the 16-bit embedding obtained from \method{} under lower precision. The \method{} column reports performance differences between \method{} and the full-precision teacher.}
% \caption{Task-type sensitivity to output embedding quantization on MMTEB (eng, v2). Each cell reports the non-negative performance drop from the corresponding 16-bit \method{} embedding under a lower embedding precision; darker green indicates larger degradation.}
\label{fig:task-type-drop}
\end{figure}

% \begin{figure}[t]
% \centering

% \begin{subfigure}[t]{0.49\linewidth}
%     \centering
%     \includegraphics[width=\linewidth]{figures/task_type_drop_heatmap.pdf}
%     \caption{Task-type sensitivity on MMTEB (eng, v2).}
%     \label{fig:task-type-drop}
% \end{subfigure}
% \hfill
% \begin{subfigure}[t]{0.49\linewidth}
%     \centering
%     \includegraphics[width=\linewidth]{figures/distillation_layer.pdf}
%     \caption{Effect of the attention-relation distillation layer selection.}
%     \label{fig:distillation-layer}
% \end{subfigure}

% \caption{
% (a) Columns 1, 2, 4, and 8 report the performance differences relative to the 16-bit embedding obtained from \method{} under lower precision. The \method{} column reports performance differences between \method{} and the full-precision teacher.
% (b) Qwen3-0.6B is evaluated on even layers from 2 to 28, while Gemma3-270M is evaluated on even layers from 2 to 18.
% }
% \label{fig:analysis-combined}
% \end{figure}

\paragraph{Multi-precision embedding performance.}
Figure~\ref{fig:precision-bar} further demonstrates the effect of output embedding precision. With multi-precision training, a single \method{} checkpoint supports multiple output embedding storage precisions and therefore offers a controllable trade-off between embedding quality and vector-index cost. 
The 8-bit and 4-bit embeddings are nearly lossless compared with the 16-bit branch on both backbones. 
% For Qwen3-0.6B, the average scores are 67.55 and 67.45, compared with 67.60 at 16-bit; for Gemma3-270M, the corresponding scores are 66.11 and 65.96, compared with 66.10 at 16-bit. 
These results show that moderate output quantization can substantially reduce storage without materially changing downstream embedding quality. Under more aggressive compression, the 1-bit and 2-bit output embeddings still preserve usable performance, achieving 64.43 and 64.90 on Qwen3-0.6B and 63.01 and 63.24 on Gemma3-270M. Although these settings introduce larger quantization noise, they reduce the storage by roughly $16\times$ and $8\times$ compared with full precision, making them attractive for storage-constrained retrieval scenarios. Overall, the multi-precision training strategy enables practitioners to select an embedding precision according to deployment constraints while retaining most of the semantic utility of the original embedding. Detailed experiments results for different embedding precisions are provided in Appendix~\ref{appendix: multi_precision}.

% Performance degradation is more pronounced at 1-bit and 2-bit output precision, especially for retrieval and reranking. These task types depend directly on fine-grained score ordering among candidate passages, and are therefore more sensitive to quantization noise in the embedding space. In contrast, classification and STS remain relatively stable across precisions, indicating that the low-bit embeddings still preserve broad semantic structure even when the vector representation is highly compressed.

% \begin{figure}[t]
% \centering
% \fbox{
% \begin{minipage}[c][1.35in][c]{0.92\linewidth}
% \centering
% MMTEB English average score vs. embedding bit-width.
% \end{minipage}}
% \caption{Performance--precision trade-off of \method{} on Qwen3-0.6B and Gemma3-270M, measured by average MMTEB English score under 1-, 2-, 4-, 8-, and 16-bit output embeddings.}
% \label{fig:precision-curve}
% \end{figure}

\subsection{Ablation Study}
\label{sec:ablation}

We conduct ablations on Qwen3-0.6B to quantify the contribution of each component. Table~\ref{tab:ablation} reports results with 16-bit output embeddings on MMTEB (eng, v2). 
As shown, removing SubLN reduces the average score from 67.60 to 65.48, showing that the additional normalization modules are important for stabilizing extreme low-bit embedding training. 
% This result supports our motivation that ternary weights and quantized activations make the backbone more sensitive to activation scale drift, and that local normalization helps preserve representation quality under quantization.
Meanwhile, distillation also plays a critical role. Removing attention-relation and similarity-distribution distillation respectively reduces the average score to 66.49 and 66.37, and removing both causes a larger drop to 64.76, indicating that teacher guidance is important for recovering the representation quality of the low-bit embedder.
% removing either distillation signal reduces performance, where using only score distillation obtains 66.49, while using only attention distillation obtains 66.37. Removing both distillation losses causes a larger drop to 64.76, showing that teacher guidance is important for recovering the representation quality of the low-bit embedder.
Continual contrastive pre-training is similarly critical. Without it, the performance of \method{} drops to 64.77, which demonstrates that continual pre-training rebuilds a broad semantic space after extreme quantization.
Finally, directly fine-tuning the converted BitNet-style backbone without continual pre-training and distillation achieves only 58.92 average score, indicating that extreme low-bit conversion substantially disrupts the embedding space and cannot be recovered by supervised fine-tuning alone.
% Direct low-bit SFT reduces the average score to 58.92. These results indicate that continual pre-training and teacher-guided distillation address complementary challenges: the former rebuilds a broad semantic space after extreme quantization, while the latter transfers fine-grained similarity and attention-structure information from the full precision teacher.

\subsection{Analysis}
\label{sec:analysis}

\paragraph{Task-type sensitivity to quantization.}
To understand the task-level sensitivity of the extreme low-bit \method{}, we compare \method{} with the full-precision teacher across different task types. As shown in the right column of Figure~\ref{fig:task-type-drop}, the \method{} remains close to the teacher on most task types, and even improves on some categories such as classification and summarization. The remaining gaps are more substantial on retrieval, which is typically harder with encoding longer documents and further embedding matching. 
Nevertheless, \method{} still maintains strong performance, suggesting that BitNet-style backbone quantization preserves most of the semantic structure needed by LLM-based embedders.
% This suggests that BitNet-style backbone quantization preserves broad semantic structure well, but ranking-oriented tasks are more sensitive to small changes in representation geometry because they depend on fine-grained ordering among many candidate passages.

Then we analyze the sensitivity introduced by quantizing output embeddings of \method{}. The 1-, 2-, 4-, and 8-bit columns in Figure~\ref{fig:task-type-drop} report differences relative to the original 16-bit embedding obtained from \method{}. Retrieval is again the most sensitive task type.
% especially at 1-bit and 2-bit precision, with changes of $-8.18$ and $-6.86$ points on Qwen3-0.6B and $-8.94$ and $-8.25$ points on Gemma3-270M. 
% Reranking and summarization also show degradation under aggressive output quantization, although the effect is smaller than retrieval. 
In contrast, classification, pair classification, and STS are more stable. 
This suggests that extreme low-bit output embeddings may mainly affect the fine-grained information required by retrieval, while still retaining sufficient semantic for tasks that rely on broader semantic separation. Overall, \method{} preserves overall semantic structure even at very low precision; with moderate low-bit quantization, it can retain performance across all task types while substantially reducing embedding storage cost.
% This indicates that low-bit output embeddings mainly affect the fine-grained similarity ordering required by retrieval, while retaining enough semantic information for tasks that rely on coarser semantic separation or pairwise consistency. Overall, \method{} preserves broad semantic structure even at very low precision, while high-recall ranking tasks benefit most from using moderate embedding precision, such as 4-bit or higher.

% \begin{figure}[t]
% \centering
% \includegraphics[width=0.7\linewidth]{figures/distillation_layer.pdf}
% \caption{Effect of the attention-relation distillation layer selection. Qwen3-0.6B is evaluated on even layers from 2 to 28, while Gemma3-270M is evaluated on even layers from 2 to 18.}
% \label{fig:distillation-layer}
% \end{figure}

% \begin{figure}[t]
% \centering
% \includegraphics[width=\linewidth]{figures/mrl_comparison.pdf}
% \caption{Comparison between \method{} and Matryoshka Representation Learning (MRL) on Qwen3-0.6B under matched storage budgets. The x-axis label $1/X$ means that the storage budget is $1/X$ of the original 1024-dimensional 16-bit output embedding. The corresponding \method{} precisions are 1, 2, 4, 8, and 16 bits, while the corresponding MRL dimensions are 64, 128, 256, 512, and 1024.}
% \label{fig:mrl-comparison}
% \end{figure}

\begin{figure}[t]
\centering
\begin{minipage}[t]{0.48\linewidth}
    \centering
    \includegraphics[width=\linewidth]{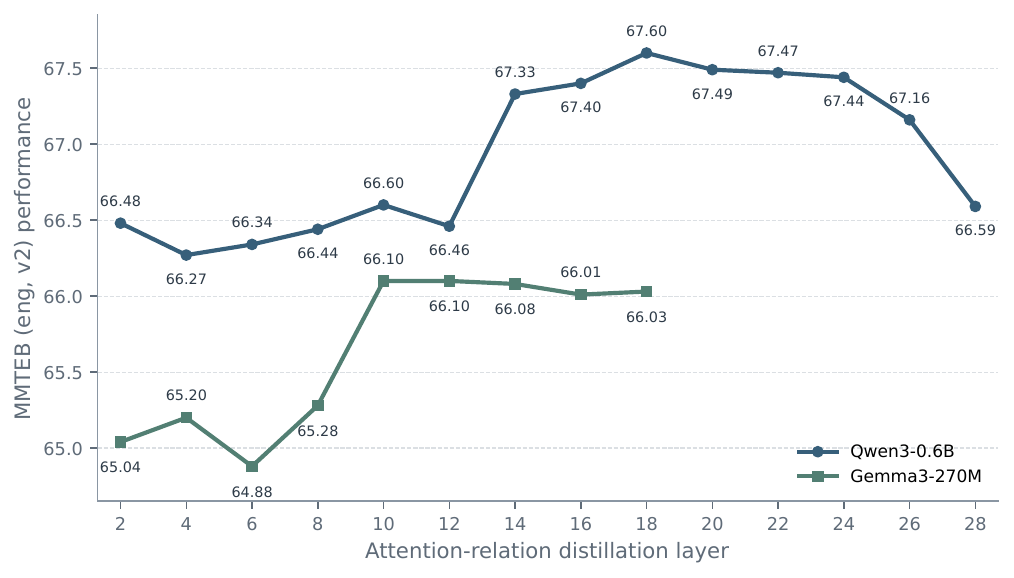}
    \caption{Effect of the attention-relation distillation layer selection. Qwen3-0.6B is evaluated on even layers from 2 to 28, while Gemma3-270M is evaluated on even layers from 2 to 18.}
    \label{fig:distillation-layer}
\end{minipage}
\hfill
\begin{minipage}[t]{0.48\linewidth}
    \centering
    \includegraphics[width=\linewidth]{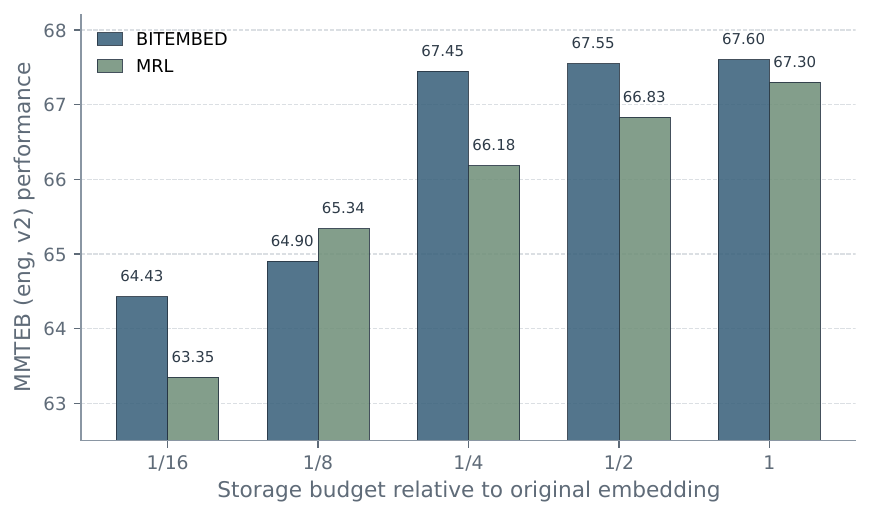}
    \caption{Comparison between \method{} and MRL. The $1/X$ means that the storage budget is $1/X$ of the original output embedding. The corresponding \method{} precisions are 1, 2, 4, 8, and 16 bits, while the corresponding MRL dimensions are 64, 128, 256, 512, and 1024.}
    \label{fig:mrl-comparison}
\end{minipage}
\end{figure}

% To understand how output embedding quantization affects different embedding tasks, we compare each low-precision setting with the 16-bit embedding of \method{} and visualize the performance drop by task type in Figure~\ref{fig:task-type-drop}. The heatmap shows a clear task-dependent pattern. Retrieval is the most sensitive task type on both backbones, especially at 1-bit and 2-bit precision, with drops of 8.18 and 6.86 points on Qwen3-0.6B and 8.94 and 8.25 points on Gemma3-270M. Reranking and summarization also show visible degradation under aggressive quantization, although the effect is smaller than retrieval. In contrast, classification, pair classification, and STS are more stable, and their drops become marginal at 4-bit and above. This pattern is consistent with the nature of dense retrieval: retrieval and reranking depend on fine-grained score ordering among many candidate passages, so small perturbations in cosine similarity can change nearest-neighbor rankings. Classification and STS, by comparison, rely more on coarse semantic separation or pairwise semantic consistency, making them less sensitive to low-bit vector noise. Overall, the analysis suggests that \method{} preserves broad semantic structure even at very low precision, while high-recall ranking tasks benefit most from using 4-bit or higher embeddings.

\paragraph{Effect of attention-relation distillation layer}
We further study the strategies for selecting the attention-relation distillation layer. Figure~\ref{fig:distillation-layer} reports the average MMTEB (eng, v2) performance when distilling attention relations from different layers. For Qwen3-0.6B, performance improves substantially from earlier layers to the middle-later layers and peaks at layer 18. For Gemma3-270M, the best results are obtained around layers 10 and 12, while earlier layers are consistently weaker. This trend is consistent across the two backbones: middle-to-later layers provide more effective distillation signals.
One possible explanation is that compared to earlier layers which mainly encode local lexical or syntactic patterns, middle-to-later layers better capture the semantic interactions that determine the final embedding space. Very late layers do not always improve performance further, suggesting that the most useful relation information for embedding distillation lies in the semantically rich but not overly specialized part of the network.

\paragraph{Effect of Multi-precision Embedding Training}
To explore the effect of multi-precision embedding training, we further compare multi-precision embedding training with a post-hoc quantization baseline that directly quantizes embeddings without exposing the model to multiple precisions during training. Table~\ref{tab:mp-analysis} reports results on Qwen3-0.6B, and detailed results are provided in Appendix~\ref{appendix:post-hoc}. Multi-precision training consistently improves performance, with larger gains at 1-, 2-, and 4-bit settings. The improvement is particularly visible on retrieval, where our method improves 1-bit and 2-bit embeddings by 2.64 and 3.10 points, respectively.
The comparison shows that by exposing \method{} to multiple embedding precisions during optimization, \method{} learns embeddings whose similarity structure is more robust. 
% This is especially beneficial for retrieval, where small score perturbations can alter the nearest-neighbor ranking.

\begin{table}[t]
\centering
\small
\caption{Effect of multi-precision embedding training on Qwen3-0.6B. ``Post'' denotes post-hoc embedding quantization without multi-precision training.}
\begin{tabular}{lcccc}
\toprule
\textbf{Bits} & \textbf{\method{}} & \textbf{Post} & \textbf{$\text{Retr.}_\text{\method{}}$} & \textbf{$\text{Retr.}_\text{Post}$} \\
\midrule
1  & 64.43 & 63.71 & 45.85 & 43.21 \\
2  & 64.90 & 64.03 & 47.17 & 44.07 \\
4  & 67.45 & 66.31 & 53.86 & 53.05 \\
8  & 67.55 & 67.29 & 54.03 & 53.58 \\
16 & 67.60 & 67.31 & 54.03 & 53.55 \\
\bottomrule
\end{tabular}
\label{tab:mp-analysis}
\end{table}

\paragraph{Comparison with Matryoshka Representation Learning}
We further compare \method{} with Matryoshka Representation Learning (MRL)~\cite{kusupati2022matryoshka}, which supports flexible storage by training nested prefix dimensions of the output embedding representation. We conduct experiments on Qwen3-0.6B, whose output embedding dimension is 1024, and we set the MRL dimensions to 64, 128, 256, 512, and 1024, matching the storage costs of the 1-, 2-, 4-, 8-, and 16-bit output embeddings of \method{} respectively for a fair comparison under the same embedding storage budgets.
As shown in Figure~\ref{fig:mrl-comparison}, \method{} achieves better performance under all other storage budgets. Although MRL is better at the 1/8 storage setting, outperforming BITEMBED by 0.44, at the 1/16, 1/4, 1/2, and full-storage settings, \method{} improves the average MMTEB score over MRL by 1.08, 1.27, 0.72, and 0.30 points.
These results suggest that compared with MRL, \method{} can keep the complete embedding space and learns to make it robust under low-bit reconstruction, allowing more semantic factors to participate in similarity computation even when the storage budget is small. Therefore, under the same storage budgets, our proposed multi-precision embedding quantization provides a more effective trade-off between embedding quality and storage efficiency.

\section{Conclusion}
In this paper, we presented \method{}, an extreme low-bit framework for LLM-based text embeddings that jointly addresses encoding efficiency and embedding storage. \method{} converts pretrained LLM backbones into BitNet-style embedders, adapting with continual contrastive pre-training, and improves representation capabilities through supervised contrastive learning with similarity-distribution and attention-relation distillation. Furthermore \method{} supports multi-precision output embeddings, enabling a single model to serve different storage budgets. Experiments on MMTEB (eng, v2) show that \method{} remains close to full-precision teachers, suggesting that extreme low-bit training is a promising direction for efficient semantic representation.

\bibliographystyle{plainnat}
\bibliography{sample-base}

\appendix
\section{Training Data}
\label{appendix: training_data}

Following BGE-en-ICL~\cite{li2024making}, our training data contains retrieval, reranking, classification, clustering, and STS tasks:

\begin{itemize}
    \item \emph{Retrieval:} ELI5\cite{fan2019eli5}, HotpotQA\cite{yang2018hotpotqa}, FEVER\cite{thorne2018fever}, MSMARCO passage and document ranking\cite{nguyen2016ms}, NQ\cite{karpukhin2020dense}, NLI\cite{gao2021simcse}, SQuAD\cite{karpukhin2020dense}, TriviaQA\cite{karpukhin2020dense}, Quora Duplicate Questions\cite{quora-question-pairs}, Arguana\cite{wachsmuth2018retrieval}, FiQA\cite{maia201818}.

\item \emph{Reranking:} SciDocsRR~\cite{cohan2020specter}, StackOverFlowDupQuestions~\cite{liu2018linkso}.

\item \emph{Classification:} AmazonReviewsClassification~\cite{mcauley2013hidden}, AmazonCounterfactualClassification~\cite{o2021wish}, Banking77Classification~\cite{casanueva2020efficient}, EmotionClassification~\cite{saravia2018carer}, TweetSentimentExtractionClassification~\cite{tweet-sentiment-extraction}, MTOPIntentClassification~\cite{li2020mtop}, IMDBClassification~\cite{maas2011learning}, ToxicConversationsClassification~\cite{enevoldsen2025mmteb}.

\item \emph{Clustering:} {Arxiv / Biorxiv / Medrxiv / Reddit / StackExchange} Clustering {S2S/P2P}, TwentyNewsgroupsClustering~\cite{lang1995newsweeder}.

\item \emph{STS:} STS12~\cite{agirre2012semeval}, STS22~\cite{chen2022semeval}, STS-Benchmark~\cite{cer2017semeval}.

\end{itemize}
\begin{table*}[!t]
\centering
\small
\footnotesize
\resizebox{0.95\linewidth}{!}{
\begin{tabular}{lp{10cm}}
\toprule
% \textbf{Models} & {{Arg}} &  {{CQG}} &  {{CQU}} & {{Cli}} & {{FEV}} & {{FiQA}} &  {{Hot}} &  {{SCI}} & {{TRE}} & {{Tou}} & {\textbf{Avg}} \\ 
Task Name &  Instruction Template \\
 \midrule

% \textbf{PRISM} & 77.64 & 64.50 & 50.75 & 41.00 & 91.89 & 57.48 & 82.08 & 26.54 & 76.56 & 61.41 & \textbf{62.98} \\
% \midrule
% LT \\
% MP \\
% MX \\
ArguAna & Given a claim, find documents that refute the claim. \\
CQADupstackGamingRetrieval & Given a question, retrieve detailed question descriptions from Stackexchange that are duplicates to the given question.\\
CQADupstackUnixRetrieval & Given a question, retrieve detailed question descriptions from Stackexchange that are duplicates to the given question. \\
ClimateFEVER & Given a claim about climate change, retrieve documents that support or refute the claim. \\
FEVERHardNegatives & Given a claim, retrieve documents that support or refute the claim \\
FiQA2018 & Given a financial question, retrieve user replies that best answer the question.\\
HotpotQAHardNegatives & Given a multi-hop question, retrieve documents that can help answer the question \\
SCIDOCS & Given a scientific paper title, retrieve paper abstracts that are cited by the given paper. \\
TRECCOVID & Given a query on COVID-19, retrieve documents that answer the query. \\
Touche2020Retrieval.v3 & Given a question, retrieve detailed and persuasive arguments that answer the question \\
STS* & Retrieve semantically similar text.\\
SummEvalSummarization.v2 & Given a news summary, retrieve other semantically similar summaries. \\
AmazonCounterfactualClassification & Classify a given Amazon customer review text as either counterfactual or not-counterfactual. \\
Banking77Classification & Given a online banking query, find the corresponding intents. \\
ImdbClassification & Classify the sentiment expressed in the given movie review text from the IMDB dataset. \\
MTOPDomainClassification & Classify the intent domain of the given utterance in task-oriented conversation. \\
MassiveIntentClassification & Given a user utterance as query, find the user intents. \\
MassiveScenarioClassification & Given a user utterance as query, find the user scenarios. \\
ToxicConversationsClassification & Classify the given comments as either toxic or not toxic. \\
TweetSentimentExtractionClassification & Classify the sentiment of a given tweet as either positive, negative, or neutral. \\
ArXivHierarchicalClusteringP2P & Identify the main and secondary category of Arxiv papers based on the titles and abstracts. \\
ArXivHierarchicalClusteringS2S & Identify the main and secondary category of Arxiv papers based on the titles.\\
BiorxivClusteringP2P.v2 & Identify the main category of Biorxiv papers based on the titles and abstracts.\\
MedrxivClusteringP2P.v2 & Identify the main category of Medrxiv papers based on the titles and abstracts.\\
MedrxivClusteringS2S.v2 & Identify the main category of Medrxiv papers based on the titles. \\
StackExchangeClustering.v2 & Identify the topic or theme of StackExchange posts based on the titles. \\
StackExchangeClusteringP2P.v2 & Identify the topic or theme of StackExchange posts based on the given paragraphs. \\
TwentyNewsgroupsClustering.v2 & Identify the topic or theme of the given news articles \\
SprintDuplicateQuestions & Retrieve duplicate questions from Sprint forum. \\
TwitterSemEval2015 & Retrieve tweets that are semantically similar to the given tweet. \\
TwitterURLCorpus & Retrieve tweets that are semantically similar to the given tweet. \\
AskUbuntuDupQuestions & Retrieve duplicate questions from AskUbuntu forum. \\
MindSmallReranking & Retrieve relevant news articles based on user browsing history. \\

\bottomrule
\end{tabular}
}
\caption{Task instructions for MMTEB.} 
\label{tab:app_instr}
\end{table*}

\begin{table*}[!t]
\centering
% \small
\footnotesize
\setlength{\tabcolsep}{1.2mm}
\resizebox{\linewidth}{!}{
\begin{tabular}{lcccc}
\toprule
\textbf{Dataset} & $\textsc{BITEMBED}_{\text{Qwen3}}$ & $\text{Teacher}_{\text{Qwen3}}$ & $\textsc{BITEMBED}_{\text{gemma3}}$ & $\text{Teacher}_{\text{gemma3}}$ \\ 
\midrule
AmazonCounterfactualClassification     & 92.19           & 92.53       & 91.55            & 91.80         \\
ArXivHierarchicalClusteringP2P         & 64.10            & 64.43       & 61.56            & 61.99        \\
ArXivHierarchicalClusteringS2S         & 59.97           & 62.10        & 58.89            & 58.97        \\
ArguAna                                & 62.81           & 66.14       & 54.77            & 64.51        \\
AskUbuntuDupQuestions                  & 58.72           & 58.32       & 56.40             & 59.09        \\
BIOSSES                                & 85.50            & 86.78       & 83.42            & 81.82        \\
Banking77Classification                & 89.13           & 88.24       & 85.25            & 89.64        \\
BiorxivClusteringP2P.v2                & 50.62           & 50.46       & 49.35            & 47.47        \\
CQADupstackGamingRetrieval             & 54.94           & 54.90        & 47.80             & 52.19        \\
CQADupstackUnixRetrieval               & 37.19           & 41.83       & 30.39            & 37.37        \\
ClimateFEVERHardNegatives              & 42.38           & 41.61       & 38.00               & 41.74        \\
FEVERHardNegatives                     & 88.78           & 89.35       & 89.41            & 88.45        \\
FiQA2018                               & 39.63           & 44.37       & 37.14            & 38.31        \\
HotpotQAHardNegatives                  & 70.84           & 73.24       & 71.71            & 73.84        \\
ImdbClassification                     & 95.07           & 95.22       & 94.79            & 94.49        \\
MTOPDomainClassification               & 97.11           & 96.84       & 95.56            & 97.60         \\
MassiveIntentClassification            & 73.90            & 73.75       & 71.52            & 73.98        \\
MassiveScenarioClassification          & 77.62           & 76.72       & 74.02            & 78.04        \\
MedrxivClusteringP2P.v2                & 47.42           & 46.79       & 44.80             & 45.04        \\
MedrxivClusteringS2S.v2                & 44.70            & 44.25       & 40.44            & 40.80         \\
MindSmallReranking                     & 28.09           & 29.47       & 27.87            & 27.69        \\
SCIDOCS                                & 21.33           & 23.28       & 20.27            & 22.00           \\
SICK-R                                 & 80.66           & 80.46       & 80.63            & 79.94        \\
STS12                                  & 78.15           & 74.39       & 77.00               & 76.75        \\
STS13                                  & 81.38           & 82.25       & 81.16            & 80.50         \\
STS14                                  & 78.60            & 79.33       & 78.70             & 78.33        \\
STS15                                  & 87.28           & 86.72       & 87.18            & 85.97        \\
STS17                                  & 89.23           & 90.01       & 89.59            & 89.99        \\
STS22.V2                               & 65.78           & 66.04       & 67.70             & 67.37        \\
STSBenchmark                           & 83.80            & 84.47       & 84.12            & 84.42        \\
SprintDuplicateQuestions               & 96.71           & 95.89       & 94.97            & 92.88        \\
StackExchangeClustering.v2             & 71.55           & 73.78       & 68.82            & 70.44        \\
StackExchangeClusteringP2P.v2          & 45.40            & 44.39       & 45.85            & 46.98        \\
SummEvalSummarization.v2               & 32.06           & 31.87       & 32.97            & 28.36        \\
TRECCOVID                              & 67.65           & 65.13       & 73.22            & 57.57        \\
Touche2020Retrieval.v3                 & 54.74           & 53.87       & 52.02            & 53.72        \\
ToxicConversationsClassification       & 89.17           & 90.62       & 91.69            & 90.69        \\
TweetSentimentExtractionClassification & 77.64           & 77.06       & 77.64            & 76.10         \\
TwentyNewsgroupsClustering.v2          & 59.57           & 57.07       & 54.83            & 59.11        \\
TwitterSemEval2015                     & 66.91           & 66.54       & 62.52            & 64.81        \\
TwitterURLCorpus                       & 84.30            & 85.26       & 83.91            & 84.30         \\ \hline
avg                                    & 67.60            & 67.95       & 66.10             & 66.71       \\

\bottomrule
\end{tabular}
}
\caption{Detail results in MMTEB (eng, v2).} 
\label{tab:mmteb_detail}
\end{table*}

\begin{table*}[t]
\centering
\small
\resizebox{\textwidth}{!}{
\begin{tabular}{lccccccccc}
\toprule
\textbf{Model} & \textbf{Bitext} & \textbf{Cls.} & \textbf{Clust.} & \textbf{PairCls.} & \textbf{Rerank} & \textbf{Retr.} & \textbf{STS} & \textbf{Avg.}\\
\midrule
\multicolumn{9}{l}{\textbf{\textit{Backbone: Qwen3-0.6B}}} \\
FP16 teacher & 82.85 & 73.88 & 54.00 & 82.07 & 63.16 & 70.15 & 77.09 & 69.00 \\ \hdashline[1pt/1pt]
$\method{}_{multilingual}$ & 81.47 & 72.65 & 53.06 & 80.47 & 62.12 & 68.33 & 74.97 & 67.49 \\
\midrule
\multicolumn{9}{l}{\textbf{\textit{Backbone: Gemma3-270M}}} \\
FP16 teacher & 81.54 & 71.84 & 52.51 & 80.12 & 61.90 & 66.38 & 75.35 & 66.55 \\ \hdashline[1pt/1pt]
$\method{}_{multilingual}$ & 80.47 & 71.09 & 52.37 & 79.72 & 60.50 & 66.71 & 74.35 & 66.26 \\
\bottomrule
\end{tabular}}
\caption{Performance of \method{} on multilingual MMTEB based on Qwen3-0.6B and Gemma3-270M.}
\label{tab:multilingual_mmteb}
\end{table*}
\begin{table*}[t]
\centering
\small
\resizebox{\textwidth}{!}{
\begin{tabular}{lccccccccc}
\toprule
\textbf{Model} & \textbf{Cls.} & \textbf{Clust.} & \textbf{PairCls.} & \textbf{Rerank} & \textbf{Retr.} & \textbf{STS} & \textbf{Summ.} & \textbf{Avg.}\\
\midrule
\multicolumn{9}{l}{\textbf{\textit{Backbone: Qwen3-0.6B}}} \\
\method{}-1bit  & 85.69 & 52.95 & 80.71 & 40.98 & 45.85 & 79.75 & 31.89 & 64.43 \\
\method{}-2bit  & 85.77 & 53.25 & 80.88 & 41.61 & 47.17 & 79.92 & 31.82 & 64.9 \\
\method{}-4bit  & 86.19 & 55.24 & 82.2  & 43.19 & 53.86 & 81.12 & 32.11 & 67.45 \\
\method{}-8bit  & 86.19 & 55.4  & 82.3  & 43.46 & 54.03 & 81.16 & 32.15 & 67.55 \\
\method{}-16bit & 86.49 & 55.42 & 82.30 & 43.41 & 54.03 & 81.15 & 32.06 & 67.60 \\
\midrule
\multicolumn{9}{l}{\textbf{\textit{Backbone: Gemma3-270M}}} \\
\method{}-1bit  & 83.83 & 51.15 & 79.36 & 40.55 & 42.53 & 81.05 & 29.84 & 63.01 \\
\method{}-2bit  & 83.91 & 51.25 & 79.64 & 41.19 & 43.22 & 80.88 & 30.30 & 63.24 \\
\method{}-4bit  & 84.99 & 52.23 & 80.91 & 42.17 & 51.11 & 81.16 & 32.55 & 65.96 \\
\method{}-8bit  & 85.26 & 53.07 & 80.47 & 42.25 & 51.50 & 81.09 & 32.96 & 66.11 \\
\method{}-16bit & 85.25 & 53.07 & 80.47 & 42.14 & 51.47 & 81.09 & 32.97 & 66.10 \\
\bottomrule
\end{tabular}}
\caption{Performance on MMTEB (eng, v2) across multi-precision output embeddings of \method{}. The ``-$b$bit" denotes the output embeddings of \method{} are stored and scored at $b$-bit precision.}
\label{tab:multi-precision}
\end{table*}
\begin{table*}[t]
\centering
\small
\resizebox{\textwidth}{!}{
\begin{tabular}{lccccccccc}
\toprule
\textbf{Model} & \textbf{Cls.} & \textbf{Clust.} & \textbf{PairCls.} & \textbf{Rerank} & \textbf{Retr.} & \textbf{STS} & \textbf{Summ.} & \textbf{Avg.}\\
\midrule
\textbf{Post}-1bit & 85.37 & 52.91 & 81.3  & 39.24 & 43.21 & 79.93 & 32.19 & 63.71 \\
\textbf{Post}-2bit & 85.52 & 53.12 & 81.26 & 39.94 & 44.07 & 80.11 & 30.63 & 64.03 \\
\textbf{Post}-4bit & 85.15 & 54.45 & 82.43 & 42.69 & 53.05 & 80.88 & 29.71 & 66.31 \\
\textbf{Post}-8bit & 86.46 & 54.86 & 82.52 & 42.73 & 53.58 & 80.94 & 30.95 & 67.29 \\
\textbf{Post}-16bit & 86.46 & 55.06 & 82.52 & 42.68 & 53.55 & 80.94 & 30.93 & 67.31 \\
\bottomrule
\end{tabular}}
\caption{Performance on MMTEB (eng, v2) across multi-precision output embeddings of post-hoc quantization baseline whose baskbone model is Qwen3-0.6B. The ``-$b$bit" denotes the output embeddings of baseline are stored and scored at $b$-bit precision.}
\label{tab:post-hoc}
\end{table*}
The total training set includes 2.1 million samples.

\section{Evaluation Details}
\label{appendix: evaluation}
% We evaluate our \method{} on MMTEB (eng, v2), which includes classification, clustering, pair classification, reranking, retrieval, semantic textual similarity (STS), and summarization tasks. The 
The instruction for evaluation on MMTEB (eng, v2) is shown in Table~\ref{tab:app_instr}.

\section{Performance on MMTEB}
\subsection{Performance Details on MMTEB (eng, v2)}
\label{appendix: overall_mmteb}
The experiments results of \method{} and teacher models based on Qwen3-0.6B and Gemma3-270M across all tasks on MMTEB (eng, v2) are available in Table~\ref{tab:mmteb_detail}.

\subsection{Performance on Multilingual MMTEB}
\label{appendix: multilingual_mmteb}
We demonstrate the performance on multilingual MMTEB of \method{} based on Qwen3-0.6B and Gemma3-270M in Table~\ref{tab:multilingual_mmteb}. In multilingual scenarios, we conduct BitNet-style quantization based on the backbone models, and conduct continual pre-training following multilingual E5~\cite{wang2024multilingual}. Then we conduct distillation-based superivised fine-tuning utilizing the traning data same as harrier-oss-v1\footnote{https://huggingface.co/microsoft/harrier-oss-v1-0.6b}.
% The experiments results of \method{} based on Qwen3-0.6B and Gemma3-270M across task categories on multilingual MMTEB are available in Table~\ref{tab:multilingual_mmteb}.

\subsection{Multi-precision embedding performance on MMTEB (eng, v2)}
\label{appendix: multi_precision}
We demonstrate the performance on MMTEB (eng, v2) of \method{} with multi-precision output embeddings in Table~\ref{tab:multi-precision}.

\subsection{Performance of post-hoc quantization baselines on MMTEB (eng, v2)}
\label{appendix:post-hoc}
We demonstrate the performance on MMTEB (eng, v2) of post-hoc quantization baseline with multi-precision output embeddings in Table~\ref{tab:post-hoc}.

\end{document}